\documentclass[letterpaper, 10 pt, conference]{ieeeconf}

\IEEEoverridecommandlockouts                             

\usepackage[english]{babel}
\usepackage{hyperref}
\usepackage{cite}
\usepackage{amsmath}
\usepackage[pdftex]{graphicx}
\usepackage{epstopdf}

\title{\LARGE \bf Fast LIDAR-based Road Detection \\ Using Fully Convolutional Neural Networks}

\author{Luca Caltagirone, Samuel Scheidegger,
	Lennart Svensson, Mattias Wahde\\
	{\tt\small\{luca.caltagirone, samsch, lennart.svensson, mattias.wahde\}@chalmers.se}
	\thanks{Luca Caltagirone and Mattias Wahde are with the Adaptive Systems Research Group, Applied Mechanics Department, Chalmers University of Technology, Gothenburg, Sweden.
Samuel Scheidegger and Lennart Svensson are with the Signal and Systems Department, also at Chalmers University of Technology. Samuel Scheidegger is also with Autoliv Research.
}}
\begin{document}

\maketitle
\thispagestyle{empty}
\pagestyle{empty}

\begin{abstract}
In this work, a deep learning approach has been developed to carry out road detection using only LIDAR data. Starting from an unstructured point cloud, top-view images encoding several basic statistics such as mean elevation and density are generated. By considering a top-view representation, road detection is reduced to a single-scale problem that can be addressed with a simple and fast fully convolutional neural network (FCN). The FCN is specifically designed for the task of pixel-wise semantic segmentation by combining a large receptive field with high-resolution feature maps. The proposed system achieved excellent performance and it is among the top-performing algorithms on the KITTI road benchmark. Its fast inference makes it particularly suitable for real-time applications.
\end{abstract}

\section{INTRODUCTION}
The estimation of free road surface (henceforth: road detection) is a crucial component for enabling fully autonomous driving \cite{HillelEtAl2014}. Besides obstacle avoidance, road detection can also facilitate path planning and decision making, especially in those situations where lane markings are not visible (for example, because covered by snow or due to poor lightning conditions) or not present (for instance, in certain rural and urban roads).
 
The problem of road detection has been investigated for many years and a large variety of approaches can be found in the literature; see, for example, \cite{HillelEtAl2014} for an in-depth survey of the field. Among the algorithms that perform best on the KITTI road benchmark data set\cite{FritschEtAl2013}, the large majority only work on monocular camera images and several make use of deep neural networks \cite{LecunEtAl2015} (DNNs). For example, in \cite{Mohan2014} the author trains deep deconvolutional networks using a multi-patch approach, while in \cite{LaddhaEtAl2016} a fully convolutional neural network (FCN) is trained with automatically annotated images. Despite achieving state-of-the-art results, camera-based approaches are strongly affected by environmental illumination. As a consequence, their performance is expected to decrease considerably at night time or whenever presented with light conditions that deviate from those seen during training.

LIDARs, on the other hand, carry out sensing by using their own emitted light and therefore they are not sensitive to environmental illumination. Road detection systems that rely on this type of sensor can then, in principle, provide the same level of accuracy across the full spectrum of light conditions experienced in daily driving, and for this reason they are particularly suitable for achieving higher levels of driving automation.
Several algorithms have been proposed that perform road detection exclusively in LIDAR point clouds or by fusing camera and LIDAR (see for example \cite{XiaoEtAl2015,HuEtAl2014,FernandesEtAl2014, ShinzatoEtAl2014}), but, to the best of our knowledge, none has used deep learning and their performance is consistently lower than the top-performing camera-based approaches.

In this paper, the problem of road detection is framed as  a \textit{pixel-wise semantic segmentation} task in point cloud top-view images using an FCN.
The proposed system carries out road segmentation in real time, on GPU-accelerated hardware, and achieves state-of-the-art performance on the KITTI road benchmark. 

The paper is organized as follows: In Section~\ref{sect:overview}, an overview and motivation of the proposed road detection system is presented and it is followed by a description of the procedure to transform an unstructured point cloud into top-view images in Section~\ref{sect:pcprepro}. The FCN's architecture is presented in Section~\ref{sect:architectures}. The data set, data augmentation, and details about the training of the FCN are described in Section~\ref{sect:data}. The results and a discussion are presented in Section~\ref{sect:experiments} and are followed by the conclusions in Section~\ref{sect:conclusion}. 

\section{POINT CLOUD TOP-VIEW ROAD DETECTION}
\label{sect:overview}
The goal of this work is to perform road detection using only LIDAR data within a deep learning framework. Here, road detection is intended as the estimation of \textit{all available} free surface for driving. Therefore, the differentiation of ego-lane versus oncoming or same-direction traffic lanes is not considered. 

Starting from an unstructured point cloud, top-view images of the vehicle's surroundings are generated. Each image encodes one of several basic statistics such as, for example, mean elevation and mean reflectivity. 
A top-view representation is, in our opinion, more appropriate than a camera perspective representation given that both path planning and vehicle control are executed in this 2D world \cite{FritschEtAl2013}. Furthermore, by using top-view images, classification is reduced to a simpler single-scale problem considering that patches of a given size cover equal surface area regardless of their position in the image. An analogous procedure for generating top-view images was also used by Chen \textit{et al.} \cite{ChenEtAl2016} but in the context of 3D object detection.

An FCN, specifically developed for semantic segmentation, is then trained to carry out road detection in the top-view images. The neural network is fully convolutional and can therefore process images of any size. An advantage of this design choice is that road detection can be carried out in regions of interest (ROIs) that can be dynamically changed and, in the case of rotating LIDARs, can even span a $360^{\circ}$ view around the vehicle.

\subsection{From point cloud to top-view images}
\label{sect:pcprepro}
An unstructured point cloud must be transformed into a suitable format before it can be used as input for an FCN. The first step of the procedure is to create a grid in the $x$\nobreakdash-$y$ plane of the LIDAR and to assign each element of the point cloud to one of its cells. The grid covers a region which is 20 meters wide, $y \in [-10, 10]$, and 40 meters long, $x \in [6,46]$, as required for the evaluation of the KITTI road benchmark; its cells are squares of size $0.10\times0.10$ meters. 

\begin{figure}[t]
	\centering
	\includegraphics[width=\columnwidth]{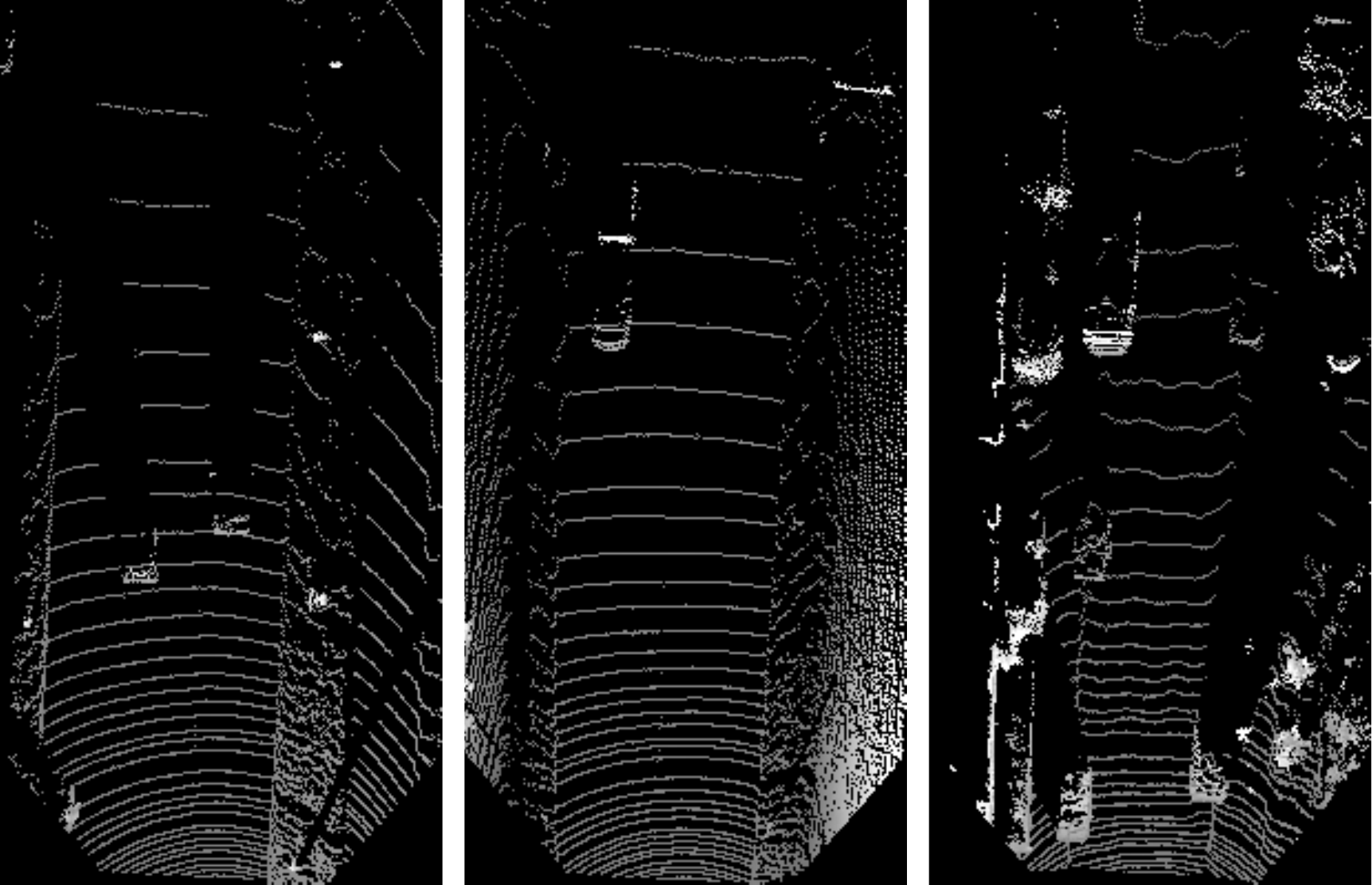}
	\caption{Three examples of mean elevation images: Each pixel grayscale value encodes the mean elevation of its relative grid cell.}
	\label{fig:meanelevation}
\end{figure}

Some basic statistics are then computed for each grid cell: number of points; mean reflectivity; as well as mean, standard deviation, minimum, and maximum elevation. 
Finally, six images, one for each of the above statistics, are generated by viewing the grid cells as pixels. Figure \ref{fig:meanelevation} shows three examples of \textit{mean elevation} images obtained with this procedure. Given the chosen cell size and grid range, these top-view images have a resolution of $200\times400$ pixels. 

\subsection{FCN architecture}
\label{sect:architectures}
In recent years, deep learning approaches have achieved state-of-the-art results in several semantic segmentation benchmarks (e.g., PASCAL VOC \cite{EveringhamEtAl2010}, MSCOCO \cite{MaireEtAl2014}, etc.).
The core idea many of those methods share is that they start from a pretrained network for image classification (e.g., VGGNet \cite{SimonyanEtAl2014}) which acts as a feature extractor, or encoder.  Additional specialized layers, such as max unpooling and deconvolution, are then added to the network in order to upsample the feature maps back to the original input size.
Some well-know deep networks that use this approach and that have inspired our FCN's architecture are Segnet \cite{BadrinarayananEtAl2015}, FCN-8s \cite{LongEtAl2015}, and Dilation \cite{YuEtAl2015}. 

\begin{figure*}[ht!]
	\centering
	\includegraphics[width=2\columnwidth]{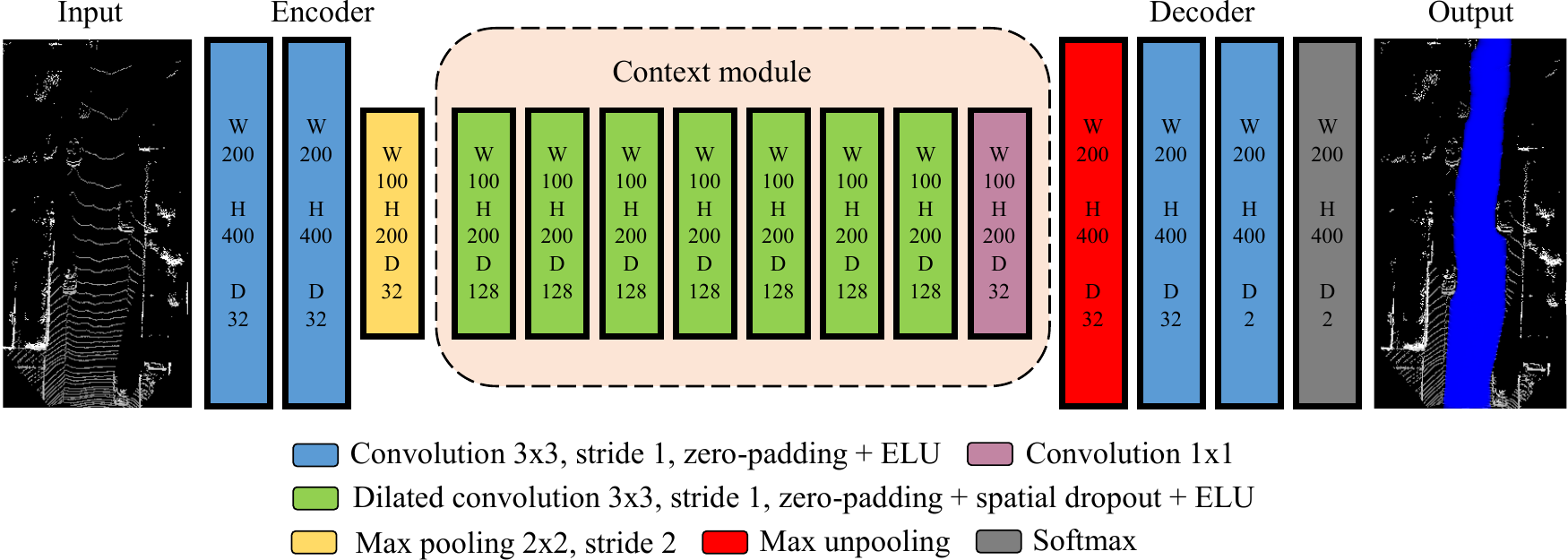}
	\caption{A schematic illustration of the proposed FCN. The input consists of 6 stacked images, one for each of the point cloud statistics described in Section \ref{sect:pcprepro}. The output is a road confidence map: The value of each pixel represents the probability that the corresponding grid cell in the LIDAR $x$-$y$ plane  belongs to the road. W represents the width, H denotes the height, and D is the number of feature maps. The FCN uses the exponential linear unit (ELU) activation function \cite{ClevertEtAl2015}.}
	\label{fig:models}
\end{figure*}

In this work, however, considering that point cloud and camera images are fundamentally different, it was deemed more appropriate to train the FCN from scratch, using only KITTI training data (see Section \ref{sect:kitti}), instead of starting from a pretrained encoder. This decision provided freedom to implement a network architecture that is specifically designed for semantic segmentation and that is tailored to the problem at hand. Particularly, the FCN was designed to have a large receptive field and to process high-resolution feature maps, two aspects that have been shown to improve segmentation accuracy \cite{WuEtAl2016}.
The FCN's architecture is shown in  Fig.~\ref{fig:models} and consists of the following components:

\begin{enumerate}
	\item A six-channel input layer, one channel for each of the point cloud statistics described in Section~\ref{sect:pcprepro}. 
	\item An encoder with the main purpose of subsampling the feature maps, thus reducing the FCN's memory requirements. Subsampling is carried out by using a max pooling layer with a $2\times2$ window and stride 2. 
	\item A context module that aggregates multi-scale contextual information by using dilated convolutions \cite{YuEtAl2015}. More details are provide in Section~\ref{sect:context_module}. 
	\item A decoder that upsamples the feature maps back to the input size by using a max-unpooling layer \cite{BadrinarayananEtAl2015} followed by two convolutional layers. 
	\item An output layer that returns a road confidence map, that is, an image where the value of each pixel represents the probability that its corresponding grid cell in the LIDAR $x$\nobreakdash-$y$ plane (see Section~\ref{sect:pcprepro}) belongs to the road. 
\end{enumerate}
 
\begin{table*}
	\centering
	\caption{Context module architecture. By using an exponentially growing dilation, the receptive field also grows exponentially without losing coverage. The feature maps are zero-padded so there is no loss of resolution. The receptive field grows at different rates in width and height, matching the input images' aspect ratio (1:2).} 
	\label{table:nets_properties}
	\begin{tabular}{|c|c|c|c|c|c|c|c|c|}
		\hline
		\multicolumn{1}{|c|}{Layer}&\multicolumn{1}{c|}{1}&\multicolumn{1}{c|}{2}&\multicolumn{1}{c|}{3}&\multicolumn{1}{c|}{4}&\multicolumn{1}{c|}{5}&\multicolumn{1}{c|}{6}&\multicolumn{1}{c|}{7} &\multicolumn{1}{c|}{8}\\
		\hline
		Filter size & $3\times3$  & $3\times3$ & $3\times3$ & $3\times3$ & $3\times3$ & $3\times3$& $3\times3$& $1\times1$\\
		Dilation (width, height) & $(1, 1)$ & $(1, 2)$ & $(2, 4)$ & $(4, 8)$ & $(8, 16)$ & $(16, 32)$& $(32, 64)$ & - \\
		Receptive field & $3\times3$  & $5\times7$ & $9\times15$ & $17\times31$ & $33\times63$ & $65\times127$& $129\times255$ & $129\times255$\\
		\# Feature maps & 128 & 128 & 128 & 128 & 128 & 128 & 128 & 32\\
		Non-linearity & ELU & ELU & ELU & ELU & ELU & ELU & ELU & -\\
		\hline
	\end{tabular}
\end{table*}

\subsection{Context module}
\label{sect:context_module}
An efficient strategy to expand the receptive field while keeping the number of parameters and layers small is to employ the dilated convolution operator that supports an exponential expansion of the receptive field without losing resolution (i.e., the feature maps do not decrease in size) or coverage \cite{YuEtAl2015}. Restricting the number of layers is important in order to reduce the FCN's memory requirements, especially when working with high-resolution feature maps.

Table~\ref{table:nets_properties} shows the implemented context module architecture; as can be seen, the receptive field of the last dilated convolution layer is larger than the input feature maps, which have a size of $100\times200$ pixels. This allows the FCN to access a large context window for inferring whether a pixel belongs to the road or not, which is particularly important considering the sparsity of point cloud top-view images (see Fig.~\ref{fig:meanelevation}).

\section{DATA SET AND SETUP}
\label{sect:data}
\subsection{The KITTI data set}
\label{sect:kitti}
The KITTI road benchmark data set \cite{FritschEtAl2013} consists of 289 training images and 290 test images taken over several days in various locations: city, rural, and highway. Ground truth annotations are represented in the camera perspective space and are only available for the training set. LIDAR point clouds are also provided as an extension to the data set. The examples are divided into three approximately equally sized (see Table \ref{table:kitti_dataset}) categories: urban unmarked (uu), urban marked (um), and urban multiple marked lanes (umm). In this work, 30 examples have been assigned to the validation set, 10 for each category. 

The training set is used for computing the objective function and adjusting the FCN's weights, while the validation set is used to decide when to stop training.
Moreover, the validation set is also used for selecting the FCN's hyper-parameters (such as, for example, number of layers, filter size, and learning rate)
by choosing the FCN (from a large set of runs) with the smallest validation error. The test set is only used for evaluating the FCN performance on unseen data, that is, its generalization error.

\begin{table}[h]
	\centering
	\caption{KITTI road dataset: size and number of images for each category and split.} 
	\label{table:kitti_dataset}
	\begin{tabular}{|c|c|c|c|c|}
		\hline
		\multicolumn{1}{|c|}{Category}&\multicolumn{1}{c|}{Train}&\multicolumn{1}{c|}{Validation}&\multicolumn{1}{c|}{Test}&\multicolumn{1}{c|}{Size [px]} \\
		\hline
		urban marked & 85  & 10 & 100 & 200 $\times$ 400 \\
		urban multiple marked & 86 & 10 & 94 & 200 $\times$ 400 \\
		urban unmarked & 88  & 10 & 100 & 200 $\times$ 400 \\
		\hline
	\end{tabular}
\end{table}

\subsection{Data augmentation}
\label{sect:data_augment}
Given that the FCNs were trained using only the KITTI data set, some simple data augmentation was necessary in order to avoid overfitting and to improve  generalization. For this purpose, each training example was rotated about the LIDAR $z$\nobreakdash-axis for angles in the range $[-30^{\circ}, 30^{\circ}]$ using steps of three degrees. 
After rotation, each example was also mirrored about the $x$\nobreakdash-axis. In this way, the data set size was increased by a factor of 42.

\subsection{Inverse perspective mapping vs. point cloud projection}
\label{sect:ann_adj}
\begin{figure}[t]
	\centering
	\includegraphics[width=\columnwidth]{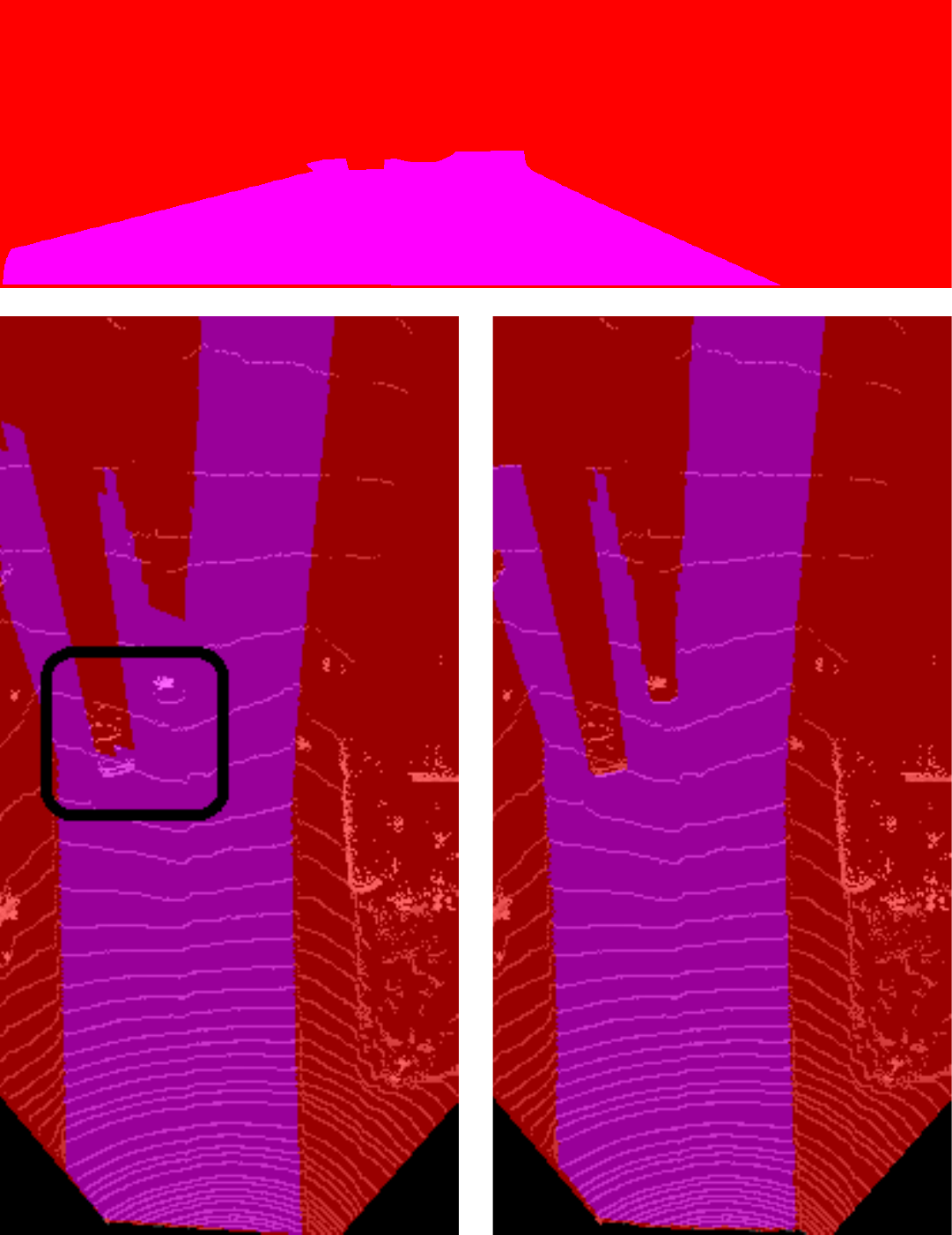}
	\caption{Top panel: Annotation in the camera perspective space. Violet and red represent road and not-road, respectively. Left panel: An example of mismatch between a point cloud top-view image and the corresponding IPM-generated annotation. The black rectangle highlights a case where the rear of a vehicle and a curb fall in the road-labeled region. Right panel: Corresponding top-view annotation obtained by projecting the interpolated point cloud onto the perspective annotation image. Here, the curb and the rear of the vehicle are both correctly marked as not-road.}
	\label{fig:mismatch}
\end{figure}

As previously mentioned,  ground truth annotations provided with the KITTI data set are represented in the camera perspective. However, given that the proposed system works with top-views of the road, the annotations must be transformed to that space for training. A possible approach to accomplish this is to use a technique known as inverse perspective mapping (IPM). Unfortunately, IPM makes the assumptions of flat and obstacle-free roads which are rarely satisfied in the real world. As a consequence, it often produces images showing inaccurate distances and road geometries. An example of this problem is illustrated in  Fig.~\ref{fig:mismatch}.

An alternative approach is to project the point cloud into the corresponding camera-view annotation in order to determine which of its points belong to the road and then use a procedure similar to the one described in Section~\ref{sect:pcprepro}, but considering the class instead of the elevation and reflectivity statistics. To increase the density of points and obtain a dense annotation, the point cloud is interpolated linearly within narrow circular sectors before carrying out the projection. The right panel of Fig.~\ref{fig:mismatch} shows an example of top-view annotation obtained by using this procedure.   

\subsection{Training}
The FCN was trained using the Adam optimization algorithm \cite{KingmaEtAl2014} with an initial learning rate of 0.01, and using cross-entropy loss as the objective function. The cross-entropy loss is defined as
\begin{equation}
L = -\frac{1}{N\times W\times H}\sum_{i=1}^{N}\sum_{m=1}^{W}\sum_{n=1}^{H} \log p^{i}_{m,n}
\end{equation}
where  $W$ and $H$ represent, respectively, the width and height of the softmax layer's output, and $N$ is the batch size which in this work was set to 4. The variable $p$ is the probability predicted by the FCN for the correct class. 
 The learning rate was decayed by a factor 2 whenever there was no improvement of performance within the last epoch. For regularization, spatial dropout layers ($p_{d}=0.25$) \cite{TompsonEtAl2015} were added after each dilated convolution layer. 
The FCN was implemented using the Torch7 framework and was trained on an NVIDIA GTX980Ti GPU with 6GB of memory. 

\section{EXPERIMENTS}
\label{sect:experiments}
\subsection{KITTI road benchmark}
The proposed road detection system was evaluated on the KITTI road benchmark test set. As mentioned in Section~\ref{sect:kitti}, this set consists of three categories: urban unmarked (uu), urban marked (um), and urban multiple marked lanes (umm). In addition, a category called \textit{urban road} is computed, which provides an overall score for these three categories combined. 
The metrics used for evaluation are precision (PRE), recall (REC), false positive rate (FPR), false negative rate (FNR), average precision (AP), and maximum F1-measure (MaxF), which is defined as follows:
\begin{equation}
\mbox{MaxF} = \max\limits_{\tau}\hspace{1.5mm} 2\times \frac{\mbox{PRE}(\tau) \times \mbox{REC}(\tau)}{\mbox{PRE}(\tau) + \mbox{REC}(\tau)},
\end{equation}
where $\tau$ is the classification threshold. 

\begin{figure}[t]
	\centering
	\includegraphics[width=\columnwidth]{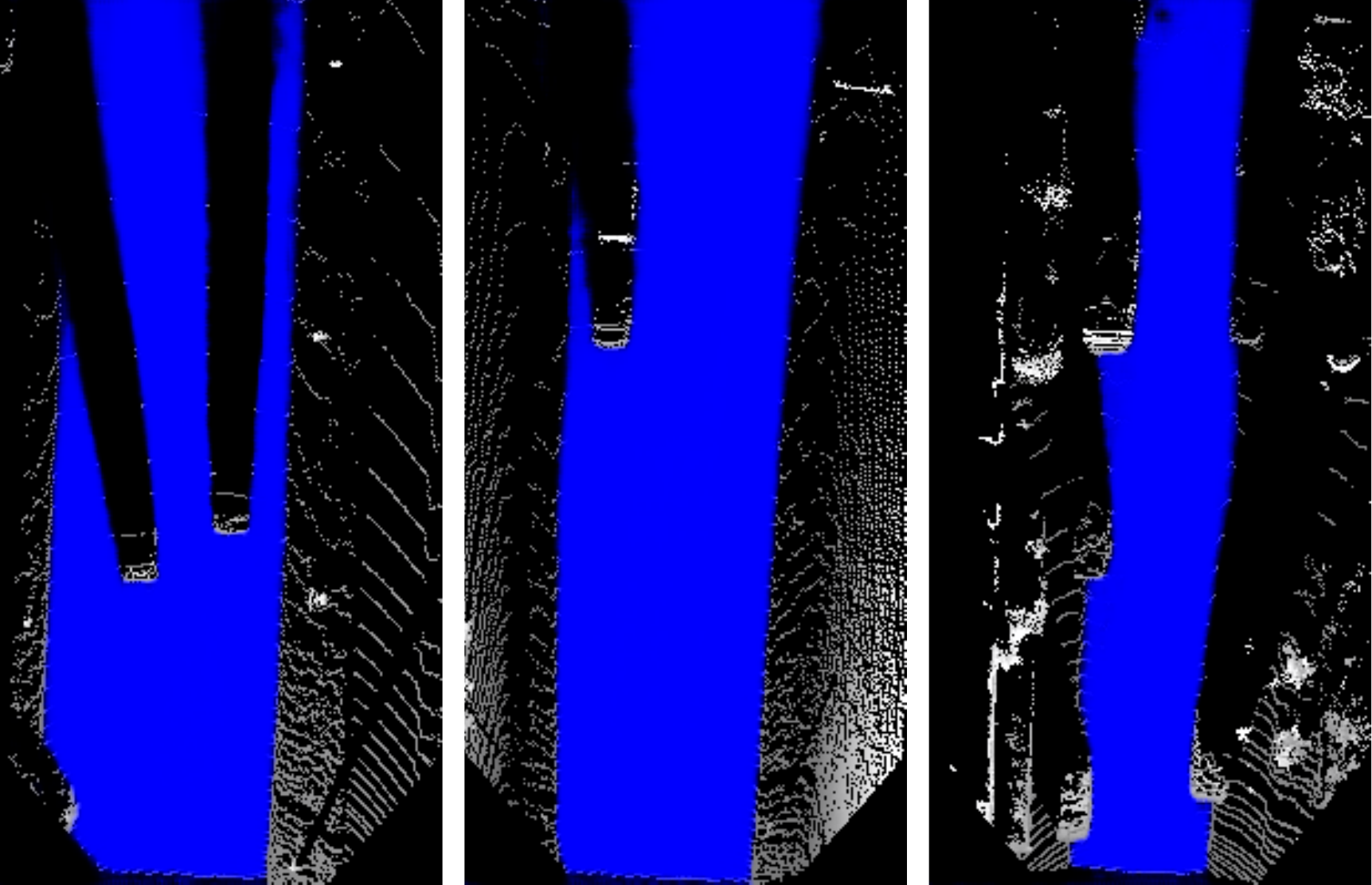}
	\caption{Road detections generated by the proposed FCN corresponding to the mean elevation images shown in Fig.~\ref{fig:meanelevation}. Higher blue intensity pixels correspond to higher probability road regions.}
	\label{fig:detectionResults}
\end{figure}

\begin{figure}[t]
	\centering
	\includegraphics[width=\columnwidth]{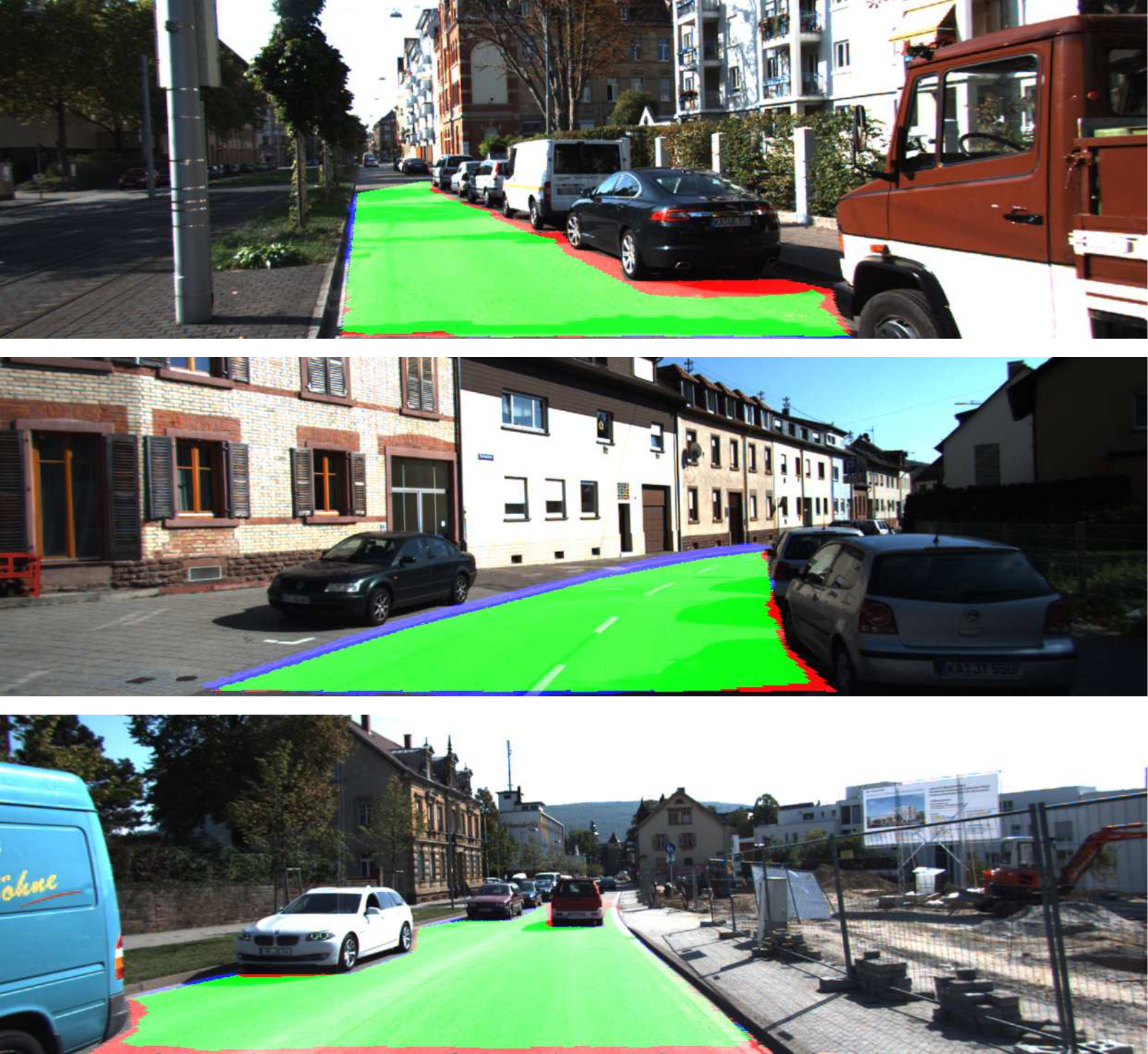}
	\caption{Examples of road detection in images from the test set. Green denotes true positive; red and blue correspond to false negative and false positive, respectively.}
	\label{fig:classificationsInPerspSpace}
\end{figure}

As shown in Table~\ref{table:kitti_results}, the proposed road detection system achieved state-of-the-art performance on the KITTI road benchmark. At the time of submission, it is the top-performing algorithm among the published methods in the overall category \textit{urban road}, outperforming by 7.4 percentage points the second best LIDAR-only system. Furthermore, its inference time is significantly smaller than most other approaches which makes it suitable for real time deployment on GPU accelerated hardware. Figure \ref{fig:detectionResults} shows some road segmentations generated by the proposed FCN on examples from the test set. As is evident from the figure, the boundary between regions that have a high probability of being part of the road and those that do not is very sharp, making the resulting road region almost uniformly blue.
The results on individual categories and additional evaluation metrics can be found at the KITTI road benchmark web page\footnote{\url{http://www.cvlibs.net/datasets/kitti/eval\_road.php}}; the proposed system is called LoDNN, which stands for LIDAR-only deep neural network.
Some examples of road segmentations projected onto the camera images are illustrated in Fig.~\ref{fig:classificationsInPerspSpace}. 
Several road detection videos can be found at \url{http://goo.gl/efLoHz}.

\begin{table}[h]
	\centering
	\caption{KITTI road benchmark results (in \%) on urban road category. Only results of published methods are reported.} 
	\label{table:kitti_results}
	\begin{tabular}{|c|c|c|c|c|c|c|}
		\hline
		{Method} & {MaxF} & {AP} & {PRE} & {REC} & {Time (ms)}\\ \hline
		\textbf{LoDNN} (our) & \textbf{94.07} & \textbf{92.03}  & 92.81 & \textbf{95.37} & \textbf{18}\\
		Up-Conv-Poly \cite{OliveiraEtAl2016} &  93.83 & 90.47  & 94.00  & 93.67 & 80\\
		DDN \cite{Mohan2014} &  93.43& 89.67  & \textbf{95.09}  & 91.82  & 2000\\
		FTP \cite{LaddhaEtAl2016} & 91.61 &90.96  & 91.04 & 92.20 & 280\\
		FCN-LC \cite{MendesEtAl2016} & 90.79 & 85.83 & 90.87 & 90.72 & 30 \\
		HIM \cite{MunozEtAl2010} & 90.64& 81.42   & 91.62  & 89.68& 7000\\
		NNP \cite{XiaozhiEtAl2015} & 89.68&  86.50  & 89.67 & 89.68 & 5000\\
		RES3D-Velo \cite{ShinzatoEtAl2014} & 86.58 & 78.34 & 82.63 & 90.92 & 360\\
		\hline
	\end{tabular}
\end{table}

\begin{figure}[t]
	\centering
	\includegraphics[width=\columnwidth]{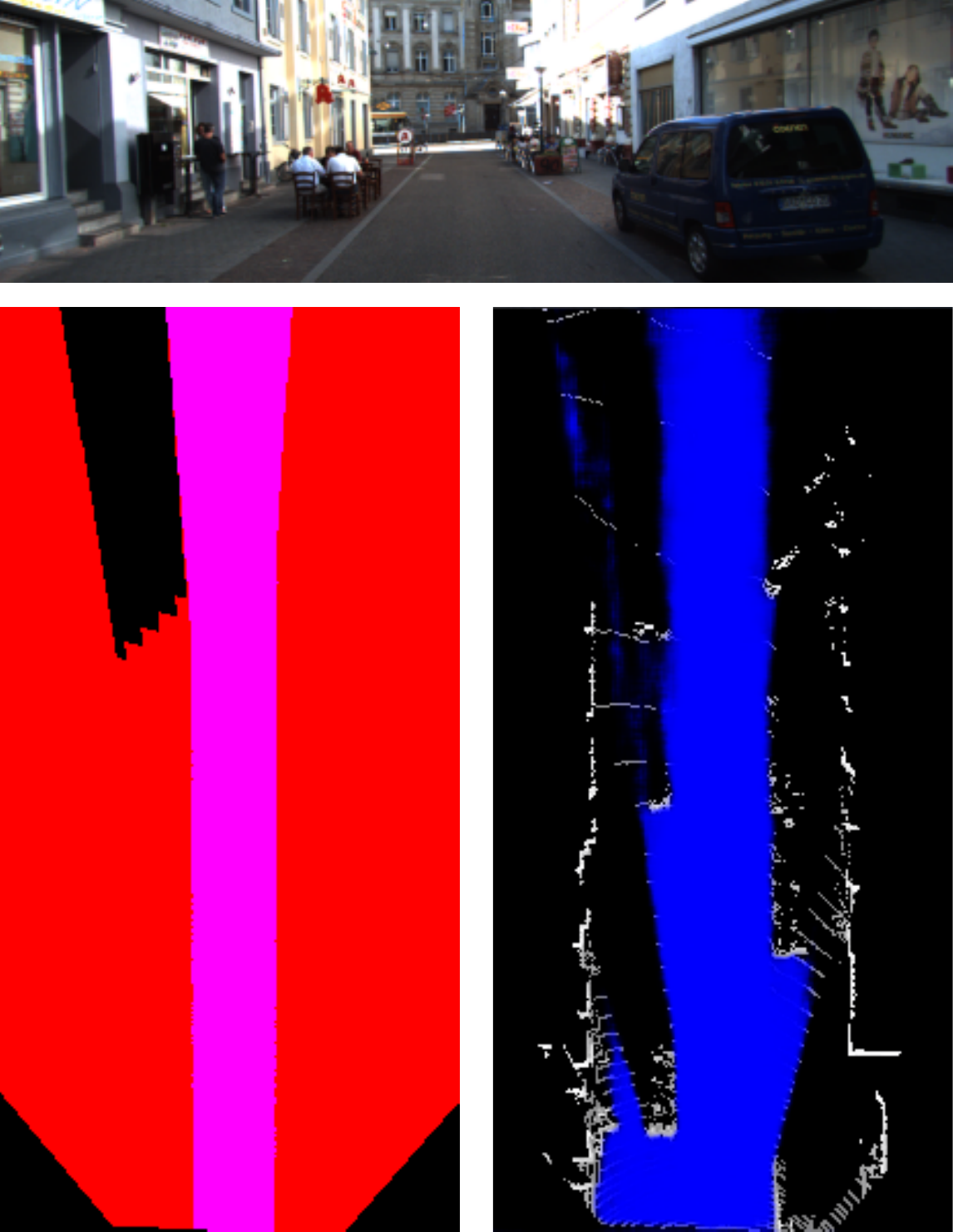}
	\caption{Top panel: Perspective camera image. Left bottom panel: Ground truth annotation. Right bottom panel: Road segmentation generated by the proposed system. The ground truth annotation labels as road only the central lane whereas the FCN's segmentation extends further to the sides which are also drivable, as illustrated by the presence of another vehicle, but that are indeed sidewalks.}
	\label{fig:urban_sidewalk}
\end{figure}

At times, false positive detections were observed in situations where the boundary between road and sidewalk was not sharp such as, for example, when the sidewalk merged with the road at pedestrian crossings, or, as illustrated in Fig.~\ref{fig:urban_sidewalk}, when the difference in elevation between road and sidewalk was very small or negligible. Complex road scenes, such as intersections, also resulted in unclear segmentations in some cases. 
These problems may be reduced by extending the training set to include more examples of such situations, using annotations explicitly made for road detection in point cloud top-views, and considering additional features for generating top-view images. 

\subsection{Regions of interest study}
LIDAR-acquired point clouds are sparse and have densities that decrease with distance from the sensor. It is therefore of relevance to evaluate how performance is affected when considering smaller ROIs with higher density of points.
According to the results shown in Table~\ref{table:smaller_regions_results}, the FCN performs best when considering regions up to 31 meters away. After that, performance degrades steadily and reaches its lowest value at the maximum considered distance of 46 meters. In order to deal with low point density in regions farther away from the LIDAR, a possible solution would be to accumulate points over successive scans. However, this is not trivial because of the presence of dynamic objects in the surroundings, such as other vehicles, as well as uncertainties introduced when estimating the ego-vehicle motion. 

\begin{table}[h]
	\centering
	\caption{ROI size study. The results pertain to the validation set. The $y$-range is [-10, 10] meters in all cases. The $x$-range lower bound is 6 meters in all cases.} 
	\label{table:smaller_regions_results}
	\begin{tabular}{|c|c|c|c|c|c|}
		\hline
		{$x$-upper bound} [m] & {MaxF} & {PRE} & {REC} & {FPR} & {FNR}\\ \hline
		46 &  95.58 & 94.15 & 97.05 & 3.33 & 2.86\\
		41 & 95.90 & 94.36 & 97.50 & 3.33 & 2.42\\
		36 & 96.13 & 94.54 & 97.78 & 3.37 & 2.15\\
		31 & 96.34 & 94.75 & 97.99 & 3.50 & 1.95\\
		26 & 96.37 & 94.78 & 98.02 & 3.80 & 1.93\\
		21 &  96.36 & 94.67 & 98.11& 4.19 & 1.92\\
		\hline
	\end{tabular}
\end{table}

\subsection{Point-cloud vs.~IPM annotations}
As explained in Section~\ref{sect:ann_adj}, top-view annotations obtained by using IPM often do not match properly with their corresponding point cloud top-view images.
Given that the evaluation for the KITTI road benchmark is carried out using IPM annotations, our system is thus penalized compared with other approaches. In order to estimate the loss of performance, a comparison has been made on the validation set using both mapping strategies. As shown in Table~\ref{table:ipmVsPC}, by using the more accurate annotations obtained by projecting the interpolated point cloud, the FCN performance increases significantly in all the considered metrics. Precision, in particular, sees an increase of 1.2 percentage points compared with the score obtained using the IPM annotations. It is therefore likely that the proposed system would achieve higher performance also on the test set if it were evaluated using more accurate top-view annotations.

\begin{table}[h]
	\centering
	\caption{Comparison IPM vs. point cloud projection (PCP) for generating top-view annotations.} 
	\label{table:ipmVsPC}
	\begin{tabular}{|c|c|c|c|c|c|c|}
		\hline
		{Split} & {Mapping} & {MaxF} & {PRE} & {REC} & {FPR} & {FNR}\\ \hline
		Validation & PCP &  95.58 & 94.15 & 97.05 & 3.33 & 2.86\\
		Validation & IPM & 94.51 & 92.92 & 96.16 & 3.94 & 3.66\\
		Test & IPM & 94.07 & 92.81 & 95.37 & 4.07 & 4.63 \\
		\hline
	\end{tabular}
\end{table}

\subsection{Occupancy images}
The statistics used for generating the point cloud top-view images were selected because they are simple and fast to compute, and only require information contained in the individual grid cells. However, it is possible to consider additional and more complex features that take into account neighboring cells or the spatial distribution of points within individual cells (e.g., principal components) and which may provide higher segmentation accuracy. Furthermore, it is far from obvious that the selected statistics and their combination are optimal for road detection. In future work, these issues may be explored more in-depth. 

An interesting preliminary result is that the FCN is able to achieve high performance: MaxF = 95.32\%, PRE = 94.15\%, REC = 96.52\%, using as input only occupancy images (i.e., white pixel if there is at least one detection, black otherwise). This indicates that the 2D distribution of points by itself, as seen from a top-view perspective, already contains strong discriminative information for road detection.  

\section{CONCLUSION}
\label{sect:conclusion}
In this work a system has been developed to perform road detection in point cloud top-view images. The proposed approach achieves state-of-the-art performance on the KITTI road benchmark, while only making use of LIDAR data, and therefore it can provide high accuracy road segmentations in any lighting conditions. Furthermore, it works in real time on GPU-accelerated hardware. Both these features make it particularly suitable for being integrated into high-level driving automation systems. 

\section*{Acknowledgment}
The authors gratefully acknowledge financial support from Vinnova/FFI. This work was partially supported by the Wallenberg Autonomous Systems and Software Program (WASP).

%% trigger a \newpage just before the given reference
%% number - used to balance the columns on the last page
%% adjust value as needed - may need to be readjusted if
%% the document is modified later
%%\IEEEtriggeratref{8}
%% The "triggered" command can be changed if desired:
%%\IEEEtriggercmd{\enlargethispage{-5in}}
%
%% references section
%
%% can use a bibliography generated by BibTeX as a .bbl file
%% BibTeX documentation can be easily obtained at:
%% http://www.ctan.org/tex-archive/biblio/bibtex/contrib/doc/
%% The IEEEtran BibTeX style support page is at:
%% http://www.michaelshell.org/tex/ieeetran/bibtex/
\bibliographystyle{IEEEtran}
%% argument is your BibTeX string definitions and bibliography database(s)
\bibliography{IEEEabrv,bibliography}

\begin{thebibliography}{10}
\providecommand{\url}[1]{#1}
\csname url@samestyle\endcsname
\providecommand{\newblock}{\relax}
\providecommand{\bibinfo}[2]{#2}
\providecommand{\BIBentrySTDinterwordspacing}{\spaceskip=0pt\relax}
\providecommand{\BIBentryALTinterwordstretchfactor}{4}
\providecommand{\BIBentryALTinterwordspacing}{\spaceskip=\fontdimen2\font plus
\BIBentryALTinterwordstretchfactor\fontdimen3\font minus
  \fontdimen4\font\relax}
\providecommand{\BIBforeignlanguage}[2]{{%
\expandafter\ifx\csname l@#1\endcsname\relax
\typeout{** WARNING: IEEEtran.bst: No hyphenation pattern has been}%
\typeout{** loaded for the language `#1'. Using the pattern for}%
\typeout{** the default language instead.}%
\else
\language=\csname l@#1\endcsname
\fi
#2}}
\providecommand{\BIBdecl}{\relax}
\BIBdecl

\bibitem{HillelEtAl2014}
A.~B. Hillel, R.~Lerner, D.~Levi, and G.~Raz, ``Recent progress in road and
  lane detection: a survey,'' \emph{Machine vision and applications}, vol.~25,
  no.~3, pp. 727--745, 2014.

\bibitem{FritschEtAl2013}
J.~Fritsch, T.~Kuehnl, and A.~Geiger, ``A new performance measure and
  evaluation benchmark for road detection algorithms,'' in \emph{International
  Conference on Intelligent Transportation Systems (ITSC)}, 2013.

\bibitem{LecunEtAl2015}
Y.~LeCun, Y.~Bengio, and G.~Hinton, ``Deep learning,'' \emph{Nature}, vol. 521,
  no. 7553, pp. 436--444, 2015.

\bibitem{Mohan2014}
R.~Mohan, ``Deep deconvolutional networks for scene parsing,'' \emph{arXiv
  preprint arXiv:1411.4101}, 2014.

\bibitem{LaddhaEtAl2016}
L.~Ankit, K.~Mehmet, S.~Luis, and M.~Hebert, ``Map-supervised road detection,''
  in \emph{IEEE Intelligent Vehicles Symposium Proceedings}, 2016.

\bibitem{XiaoEtAl2015}
L.~Xiao, B.~Dai, D.~Liu, T.~Hu, and T.~Wu, ``Crf based road detection with
  multi-sensor fusion,'' in \emph{Intelligent Vehicles Symposium (IV)}, 2015.

\bibitem{HuEtAl2014}
X.~Hu, F.~S.~A. Rodriguez, and A.~Gepperth, ``A multi-modal system for road
  detection and segmentation,'' in \emph{2014 IEEE Intelligent Vehicles
  Symposium Proceedings}.\hskip 1em plus 0.5em minus 0.4em\relax IEEE, 2014,
  pp. 1365--1370.

\bibitem{FernandesEtAl2014}
R.~Fernandes, C.~Premebida, P.~Peixoto, D.~Wolf, and U.~Nunes, ``Road detection
  using high resolution lidar,'' in \emph{2014 IEEE Vehicle Power and
  Propulsion Conference (VPPC)}, Oct 2014, pp. 1--6.

\bibitem{ShinzatoEtAl2014}
P.~Y. Shinzato, D.~F. Wolf, and C.~Stiller, ``Road terrain detection: Avoiding
  common obstacle detection assumptions using sensor fusion,'' in \emph{2014
  IEEE Intelligent Vehicles Symposium Proceedings}.\hskip 1em plus 0.5em minus
  0.4em\relax IEEE, 2014, pp. 687--692.

\bibitem{ChenEtAl2016}
X.~Chen, H.~Ma, J.~Wan, B.~Li, and T.~Xia, ``Multi-view 3d object detection
  network for autonomous driving,'' \emph{arXiv preprint arXiv:1611.07759},
  2016.

\bibitem{EveringhamEtAl2010}
\BIBentryALTinterwordspacing
M.~Everingham, L.~Van~Gool, C.~K.~I. Williams, J.~Winn, and A.~Zisserman, ``The
  pascal visual object classes (voc) challenge,'' \emph{International Journal
  of Computer Vision}, vol.~88, no.~2, pp. 303--338, 2010. [Online]. Available:
  \url{http://dx.doi.org/10.1007/s11263-009-0275-4}
\BIBentrySTDinterwordspacing

\bibitem{MaireEtAl2014}
\BIBentryALTinterwordspacing
T.~Lin, M.~Maire, S.~J. Belongie, L.~D. Bourdev, R.~B. Girshick, J.~Hays,
  P.~Perona, D.~Ramanan, P.~Doll{\'{a}}r, and C.~L. Zitnick, ``Microsoft
  {COCO:} common objects in context,'' \emph{CoRR}, vol. abs/1405.0312, 2014.
  [Online]. Available: \url{http://arxiv.org/abs/1405.0312}
\BIBentrySTDinterwordspacing

\bibitem{SimonyanEtAl2014}
K.~Simonyan and A.~Zisserman, ``Very deep convolutional networks for
  large-scale image recognition,'' \emph{arXiv preprint arXiv:1409.1556}, 2014.

\bibitem{BadrinarayananEtAl2015}
V.~Badrinarayanan, A.~Handa, and R.~Cipolla, ``Segnet: A deep convolutional
  encoder-decoder architecture for robust semantic pixel-wise labelling,''
  \emph{arXiv preprint arXiv:1505.07293}, 2015.

\bibitem{LongEtAl2015}
J.~Long, E.~Shelhamer, and T.~Darrell, ``Fully convolutional networks for
  semantic segmentation,'' in \emph{Proceedings of the IEEE Conference on
  Computer Vision and Pattern Recognition}, 2015, pp. 3431--3440.

\bibitem{YuEtAl2015}
F.~Yu and V.~Koltun, ``Multi-scale context aggregation by dilated
  convolutions,'' \emph{arXiv preprint arXiv:1511.07122}, 2015.

\bibitem{WuEtAl2016}
\BIBentryALTinterwordspacing
Z.~Wu, C.~Shen, and A.~van~den Hengel, ``High-performance semantic segmentation
  using very deep fully convolutional networks,'' \emph{CoRR}, vol.
  abs/1604.04339, 2016. [Online]. Available:
  \url{http://arxiv.org/abs/1604.04339}
\BIBentrySTDinterwordspacing

\bibitem{ClevertEtAl2015}
D.-A. Clevert, T.~Unterthiner, and S.~Hochreiter, ``Fast and accurate deep
  network learning by exponential linear units (elus),'' \emph{arXiv preprint
  arXiv:1511.07289}, 2015.

\bibitem{KingmaEtAl2014}
D.~Kingma and J.~Ba, ``Adam: A method for stochastic optimization,''
  \emph{arXiv preprint arXiv:1412.6980}, 2014.

\bibitem{TompsonEtAl2015}
J.~Tompson, R.~Goroshin, A.~Jain, Y.~LeCun, and C.~Bregler, ``Efficient object
  localization using convolutional networks,'' in \emph{Proceedings of the IEEE
  Conference on Computer Vision and Pattern Recognition}, 2015, pp. 648--656.

\bibitem{OliveiraEtAl2016}
G.~L. Oliveira, W.~Burgard, and T.~Brox, ``Efficient deep methods for monocular
  road segmentation.'' in \emph{IEEE/RSJ International Conference on
  Intelligent Robots and Systems (IROS 2016)}, 2016.

\bibitem{MendesEtAl2016}
C.~C.~T. Mendes, V.~Fr{\'{e}}mont, and D.~F. Wolf, ``Exploiting fully
  convolutional neural networks for fast road detection,'' in \emph{2016 IEEE
  International Conference on Robotics and Automation (ICRA)}, May 2016, pp.
  3174--3179.

\bibitem{MunozEtAl2010}
D.~Munoz, J.~A. Bagnell, and M.~Hebert, ``Stacked hierarchical labeling,'' in
  \emph{European Conference on Computer Vision (ECCV)}, 2010.

\bibitem{XiaozhiEtAl2015}
X.~Chen, K.~Kundu, Y.~Zhu, A.~Berneshawi, H.~Ma, S.~Fidler, and R.~Urtasun,
  ``3d object proposals for accurate object class detection,'' in \emph{NIPS},
  2015.

\end{thebibliography}

\end{document}